\documentclass{article}
\usepackage{ijcai22}

\usepackage{times}
\usepackage{soul}
\usepackage{url}
\usepackage[hidelinks]{hyperref}
\usepackage[utf8]{inputenc}
\usepackage[small]{caption}
\usepackage{graphicx}
\usepackage{amsmath}
\usepackage{amsthm}
\usepackage{amssymb}
\usepackage{booktabs}
\usepackage{algorithm}
\usepackage{algorithmic}
\usepackage{svg}
\usepackage{todonotes}
\usepackage{natbib}
\renewcommand{\S} {\mathcal{S}}
\newcommand{\A} {\mathcal{A}}
\newcommand{\G} {\mathcal{G}}
\newcommand{\B} {\mathcal{B}}

\newcommand{\bc} {\mathbf{b}}
\newcommand{\E} {\mathbb{E}}
\newcommand{\R} {\mathcal{R}}
\newcommand{\T}{\mathcal{T}}
\newcommand{\M} {\mathcal{M}}
\newcommand{\0} {\mathbf{0}}
\newcommand{\nov} {\operatorname{novel}}
\renewcommand{\P} {\mathbb{P}}

\DeclareMathOperator*{\argmax}{argmax}

\title{Scaling Goal-based Exploration via Pruning Proto-goals}

\author{}

\author{
Akhil Bagaria$^1$\footnote{Work done during an internship at DeepMind, London.} 
\And Ray Jiang$^2$
\And Ramana Kumar$^2$
\And Tom Schaul$^2$
\affiliations
$^1$Brown University, Providence, RI, USA\\
$^2$DeepMind, London, UK
\emails
\texttt{akhil\_bagaria@brown.edu},\,\, \texttt{tom@deepmind.com}
}

\begin{document}

\maketitle

\begin{abstract}
One of the gnarliest challenges in reinforcement learning (RL) is exploration that scales to vast domains, where novelty-, or coverage-seeking behaviour falls short. Goal-directed, purposeful behaviours are able to overcome this, but rely on a good goal space. The core challenge in \emph{goal discovery} is finding the right balance between generality (not hand-crafted) and tractability (useful, not too many). Our approach explicitly seeks the middle ground, enabling the human designer to specify a vast but meaningful \emph{proto-goal} space, and an autonomous discovery process to refine this to a narrower space of controllable, reachable, novel, and relevant goals. The effectiveness of goal-conditioned exploration with the latter is then demonstrated in three challenging environments.
\end{abstract}

\section{Introduction} 
\label{sec:introduction}

Exploration is widely recognised as a core challenge in RL. It is most acutely felt when scaling to vast domains, where classical novelty-seeking methods are insufficient \citep{taiga2021bonus} because there are simply too many things to observe, do, and learn about; and the agent's lifetime is far too short to approach exhaustive coverage \citep{sutton2022alberta}.

Abstraction can overcome this issue \citep{gershman2017blessing,konidaris2019necessity}: by learning about goal-directed, purposeful behaviours (and how to combine them), the RL agent can ignore irrelevant details, and effectively traverse the state space.
Goal-conditioned RL is one natural formalism of abstraction, and especially appealing when the agent can learn to generalise across goals \citep{uvfaSchaul15}.

The effectiveness of goal-conditioned agents directly depends on the size and quality of the goal space (Section~\ref{sec:protogoals}). If it is too large, such as treating all encountered states as goals \citep{andrychowicz2017neurips}, most of the abstraction benefits vanish.
On the other extreme, hand-crafting a small number of useful goals \citep{barreto2019option} limits the generality of the method.
The answer to this conundrum is to adaptively expand or refine the goal space based on experience, also known as the \emph{discovery} problem, allowing for a more autonomous agent that can be both general and scalable.

\begin{figure}[tb]
    \centering
    \vspace{-1.5em}
    \includegraphics[width=0.7\linewidth]{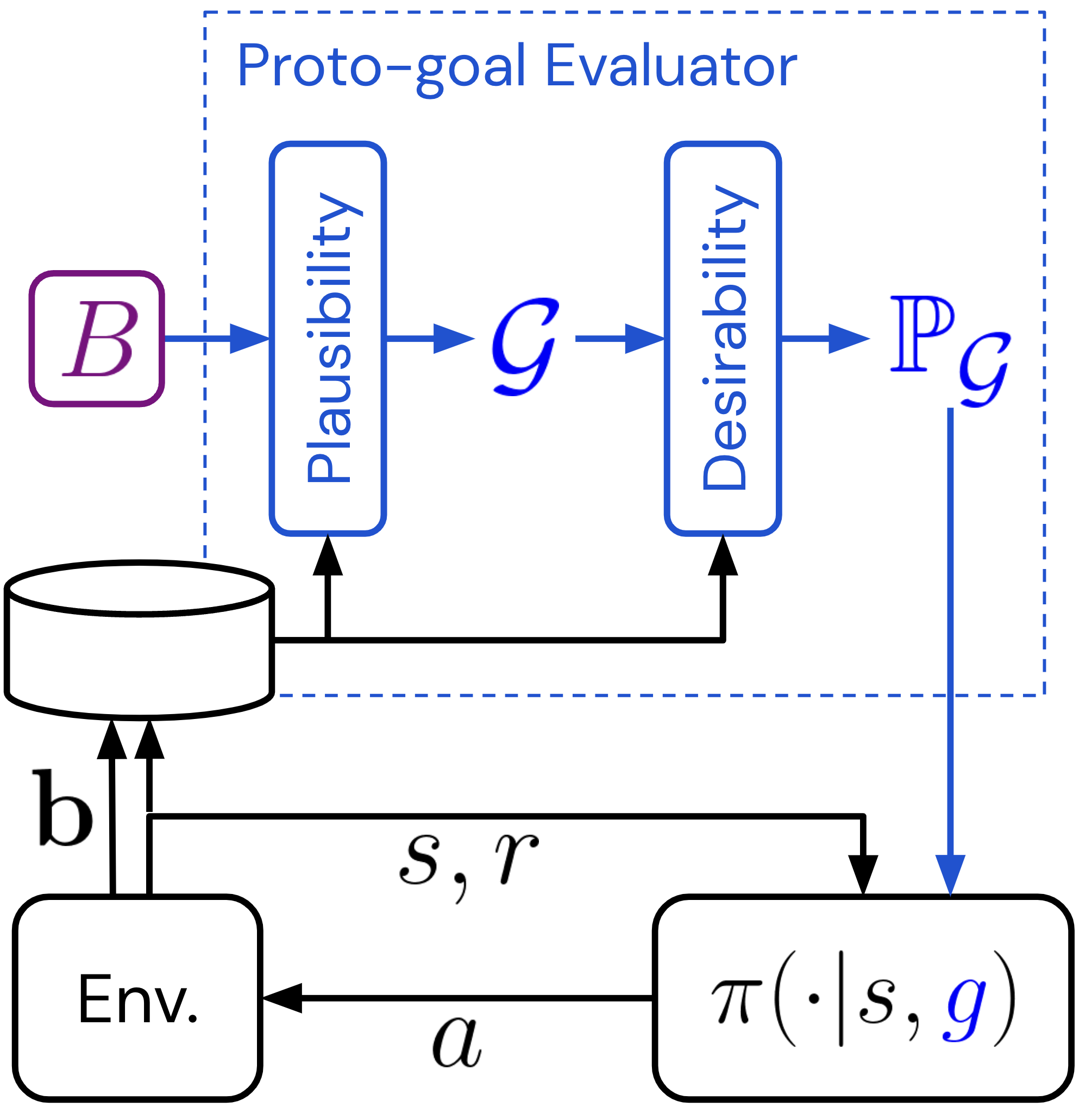}
    \vspace{-0.5em}
    \caption{{\bf Proto-goal RL}: a goal-conditioned RL agent's policy $\pi$ acts with goals $g$ obtained from its Proto-goal Evaluator (PGE, {\color{blue} blue}). The PGE refines a cheaply defined \textit{proto-goal space} ($B$, {\color{violet}violet}) into a smaller set of plausible goals $\G$, using observed transition data $(s,a,r,s')$ that includes information about encountered proto-goals ($\bc$). 
    It further endows $\G$ with a distribution $\P_\G$, based on goal desirability, from which $g$ is then sampled.
    }
    \vspace{-0.5em}
    \label{fig:proto_goal_rl}
\end{figure}

Taking a step towards this ultimate aim, we propose a framework with two elements.
First, a \emph{proto-goal} space (Section~\ref{sec:protogoals}), which can be cheaply designed to be meaningful for the domain at hand, e.g., by pointing out the most salient part of an observation using  domain knowledge \citep{chentanez2004intrinsically}.
What makes defining a proto-goal space much easier than defining a goal space is its leniency: it can remain (combinatorially) large and unrefined, with many uninteresting or useless proto-goals.
Second, an adaptive function mapping this space to a compact set of useful goals, 
called a \emph{Proto-goal Evaluator} (PGE, Section~\ref{sec:pge}).
The PGE may employ multiple criteria of usefulness, such as controllability, novelty, reachability, learning progress, or reward-relevance.
Finally we address pragmatic concerns on how to integrate these elements into a large-scale goal-conditioned RL agent (Section~\ref{sec:gcrl}), and show it can produce a qualitative leap in performance in otherwise intractable exploration domains (Section~\ref{sec:experiments}).

\section{Background and Related Work}
We consider problems modeled as Markov Decision Processes (MDPs) $\M=(\S,\A,\R,\T,\gamma)$, where $\S$ is the state space, $\A$ is the action space, $\R$ is the reward function, $\T$ is the transition function and $\gamma$ is the discount factor. The aim of the agent is to learn a policy that maximises the sum of expected rewards \citep{sutton2018reinforcement}.

\paragraph{Exploration in RL.} 
Many RL systems use dithering strategies for exploration (e.g., $\epsilon$-greedy, softmax, action-noise \citep{lillicrap2015continuous} and parameter noise \citep{fortunato2017noisy,plappert2017parameter}). Among those that address \textit{deep} exploration, the majority of research \citep{taiga2021bonus} has focused on count-based exploration \citep{strehl2008analysis,bellemare2016unifying}, minimizing model prediction error \citep{pathak2017curiosity,burdaLargeScaleStudy2019,burda2018exploration}, or picking actions to reduce uncertainty \citep{osband2016deep,osband2018randomized} over the state space. 
These strategies try to eventually learn about \textit{all} states, which might not be a scalable strategy when the world is a lot bigger than the agent \citep{sutton2022alberta}.
We build on the relatively under-studied family of exploration methods that maximize the agent's \textit{learning progress} \citep{schmidhuber1991curious,kaplan2004maximizing}.

\paragraph{General Value Functions.} 
Rather than being limited to predicting and maximizing a single reward (as in vanilla RL), General Value Functions (GVFs) \citep{sutton2011horde} predict (and sometimes control \citep{jaderberg2016reinforcement}) ``cumulants'' that can be constructed out of the agent's sensorimotor stream. The discounted sum of these cumulants are GVFs and can serve as the basis of representing rich knowledge about the world \citep{schaul2013better,veeriah2019discovery}. 

\paragraph{Goal-conditioned RL.} 
When the space of cumulants is limited to goals, GVFs reduce to goal-conditioned value functions that are often represented using Universal Value Function Approximators (UVFAs) \citep{uvfaSchaul15}. Hindsight Experience Replay (HER) is a popular way of learning UVFAs in a sample-efficient way \citep{andrychowicz2017neurips}. The two most common approaches is to assume that a set of goals is given, or to treats all observations as potential goals \citep{liu2022goal} and try to learn a controller that can reach \textit{any} state. In large environments, the latter methods often over-explore \citep{pong2019skew,pitis2020maximum} or suffer from interference between goals \citep{schaul2019ray}.

\paragraph{Discovery of goals and options.} 
Rather than assuming that useful goals are pre-specified by a designer, general-purpose agents must \textit{discover} their own goals or options \citep{sutton1999between}. Several heuristics have been proposed for discovery (see \citet{daveabel2020abstractions} Ch 2.3 for a survey): reward relevance \citep{bacon2017option,veeriah2021discovery}, composability \citep{konidaris2009skill,bagaria2019option}, diversity \citep{eysenbach2018diversity,campos2020explore}, empowerment \citep{mohamed2015variational}, coverage \citep{bagaria2021skill,machado2017laplacian}, etc. These heuristics measure \textit{desirability}, but they must be paired with \textit{plausibility} metrics like controllability and reachability to discover meaningful goals in large goal spaces. The IMGEP framework \citep{forestier2022intrinsically} also does skill-acquisition based on competence progress, but they assume more structure in the goal space (e.g., Euclidean measure, objects), and use evolution strategies to represent policies instead of RL.


\section{Goals and Proto-goals}
\label{sec:protogoals}

A \emph{goal} is anything that an agent can pursue and attain through its behaviour. Goals are well formalised with a scalar cumulant $c_g: \S\times\A\times\S \rightarrow \mathbb{R}$ and a continuation function $\gamma_g: \S\times\A\times\S \rightarrow [0,1]$, as proposed in the general value function (GVF) framework \citep{sutton2011horde}. Here, we consider the subclass of \emph{attainment} goals $g$, or ``endgoals'', which imply a binary reward that is paired with termination. In other words a transition has either $(c_g=0, \gamma_g>0)$ or $(c_g=1, \gamma_g=0)$, i.e., only terminal transitions are rewarding.
The corresponding goal-optimal value functions satisfy:
\[
Q^{*}_g(s,a) = \E_{s'} \left[
c_g(s,a,s') +  \gamma_g(s,a,s') \max_{a'} Q^{*}_g(s',a')
\right],
\]
with corresponding greedy policy $\pi^{*}_g := \arg\max_a Q^{*}_g(s,a)$.

\emph{Proto-goals} are sources of goals. Since attainment goals can easily be derived from any binary function, we formally define a proto-goal to be a binary function of a transition $b_i: \S\times\A\times\S \rightarrow \{0,1\}$.
We assume that, for a given domain, a set $B$ of such proto-goals can be queried.
Proto-goals differ from goals in two ways. First, to fully specify a goal, a proto-goal must be paired with a time-scale constant $\gamma \in [0,1]$ (a discount), which defines the horizon over which $g$ should be achieved.
The pair $(b_i,\gamma)$ then define the goal's cumulant $c_g(s,a,s') := b_i(s,a,s')$ and continuation function $\gamma_g(s,a,s') := \gamma (1-b_i(s,a,s'))$.
Second, less formally, the space of proto-goals $B$ is vastly larger than any reasonable set goals $\G$ that could be useful to an RL agent. Hence the need for the Proto-goal evaluator (Section~\ref{sec:pge}) to convert one space into the other.

\subsection{Example Proto-goal Spaces}
A proto-goal space implicitly defines a large, discrete space of goals. Its design uses some domain knowledge, but, crucially, no direct knowledge of how to reach the solution.
The most common form is to use designer knowledge about which aspects of an observation are most salient. For example, many games have on-screen counters that track task-relevant quantities (health, resources, etc.). Other examples include treating inventory as privileged in \textsc{Minecraft}, sound effects in console video games, text feedback in domains like \textsc{NetHack} (see Section~\ref{sec:minihack} and Figure~\ref{fig:minihack_proto_illustration}), or object attributes in robotics.
In all of these cases, it is straightforward to build a set of binary functions---for example, in \textsc{NetHack}, a naive proto-goal space includes one binary function for each possible word that could be present in the text feedback.

\subsection{Representation}
Each observation from the environment is accompanied by a binary proto-goal vector $\bc_t \in \{0,1\}^{|B|}$, with entries of $1$ indicating which proto-goals are achieved in the current state (Figure~\ref{fig:proto_goal_rl}).
Initially, the agent decomposes $\bc_t$ into $1$-hot vectors, focusing on goals that depend on a single dimension. As the agent begins to master $1$-hot goals, it combines them using the procedure described in Section~\ref{sec:recombination}, to expand the goal space and construct multi-hot goals.

When querying the goal-conditioned policy $\pi(a|s,g)$, we use the same $1$-hot or multi-hot binary vector representation for the goal $g$.

\subsection{Goal Recombinations} 
\label{sec:recombination}

A neat side-effect of a binary proto-goal space $B$ is that it can straightforwardly be extended to a combinatorially larger goal space with logical operations.
For example, using the logical \texttt{AND} operation, we can create goals that are only attained once multiple separate bits of $\bc$ are activated simultaneously.\footnote{Note that we combine goals, but not their corresponding value-functions \citep{barreto2019option,tasse2022world}; we let the UVFA $Q_{\theta}(s, a, g)$ handle generalization to newly created goals and leave combination in value-function space to future work.}
One advantage of this is that it places less burden on the design of the proto-goal space, because $B$ only needs to contain useful goal components, not the useful goals themselves. This is also a form of continual learning \citep{ring1994continual}, with more complex or harder-to-reach goals continually being constructed out of existing ones.
The guiding principle to keep this constructivist process from drowning in too many combinations is to operate in a gradual fashion: we only combine goals that in addition to being plausible and desirable (Section~\ref{sec:pge}), have also been \emph{mastered} (Section~\ref{sec:mastery}).

\section{Proto-goal Evaluator} 
\label{sec:pge}

The Proto-goal Evaluator (PGE) converts the large set of  proto-goals to a smaller, more interesting set of goals $\G$. 
It does this in two stages: a binary filtering stage that \emph{prunes} goals by plausibility, and a weighting stage that creates a \emph{distribution} over the remaining goals $\P_{\G}: \G \rightarrow [0,1]$, based on desirability.

\subsection{Plausibility Pruning} 
\label{sec:plausibility}

Implausible proto-goals are those that most likely cannot be achieved (either \textit{ever} or given the current data distribution). Having them in the goal space is unlikely to increase the agent's competence; to the contrary, they can distract and hog capacity. We use the following three criteria to eliminate implausible goals:
\begin{description}
    \item[Observed:] we prune any proto-goal $b_i$ that has never been observed in the agent's experience, so far.
    \item[Reachable:] we prune proto-goals that are deemed unreachable (e.g., pigs cannot fly, a person cannot be in London and Paris at the same time).
    \item[Controllable:] similarly, we prune goals that are outside of the agent's control (e.g., sunny weather is reachable, but not controllable).
\end{description}

For the first criterion, we simply track global counts $N(g)$ for how often we have observed the proto-goal $b_i$ that corresponds to $g$ being reached.
Estimating reachability and controllability is a bit more involved.
We do this by computing a pair of \emph{proxy} value functions:
each goal $g$ is associated with two types of reward functions (or cumulants)---one with ``seek'' semantics and the other with ``avoid'' semantics:
\begin{align*}
    R_{\text{seek}}(s, g) &= 1 \text{ if } g \text{ is achieved in } s \text{ else } 0 \\
    R_{\text{avoid}}(s, g) &= -1 \text{ if } g \text{ is achieved in } s \text{ else } 0.
\end{align*}
These seek/avoid cumulants in turn induce seek/avoid policies, and value functions $V_{\text{seek}}, V_{\text{avoid}}$ that correspond to these policies. Estimates of these values are learned from transitions stored in the replay buffer $\B$.

A proto-goal $g$ is {\bf globally reachable} if it can be achieved from \textit{some} state: 
\begin{equation}
\label{eq:reach}
\max_{s \sim \B} V_{\text{seek}}(s, g) > \tau_1,
\end{equation}
where $\tau_1>0$ is a threshold representing the (discounted) probability below which a goal is deemed to be unreachable.

A proto-goal $g$ is judged as {\bf uncontrollable} if a policy seeking it is equally likely to encounter it as a policy avoiding it: 
\begin{equation}
\E_s\Big[V_{\text{seek}}(s, g)\Big] - \E_s\Big[-V_{\text{avoid}}(s, g)\Big] < \tau_2,
\label{eq:control}
\end{equation}
up to threshold $\tau_2$.
The set of plausible goals $\G$ is the subset of those proto-goals induced by $B$ that satisfy both Eq.~\ref{eq:reach} and~\ref{eq:control}.

\subsubsection{Scalably Estimating Many Seek/Avoid Values with LSPI}
\label{sec:seek_avoid_lspi}

As a first line of defense in the process of trimming a vast proto-goal space, the reachability and controllability estimation (and hence the computation of the proxy values $V_{\text{seek}}, V_{\text{avoid}}$) must be very cheap per goal considered.
On the other hand, their accuracy requirement is low: they are not used for acting or updating other values via TD \citep{sutton1988learning}, and it suffices to eliminate \textit{some} fraction of implausible goals.
Consequently, we have adopted four radical simplifications that reduce the compute requirements of estimating proxy values, to far less than is used in the main deep RL agent training.
First, we reduce the value estimation to a \emph{linear} function approximation problem, by invoking two iterations of least-squares policy iteration (LSPI, \citep{lagoudakis2003least,ghavamzadeh2010lstd}), one for the ``seek'' and one for the ``avoid'' policy.
As input features for LSPI we use random projections of the observations into $\mathbb{R}^{|\phi|}$, which has the added benefit of making this approach scalable independently of the observation size.
Third, the estimation is done on a batch of transitions that are only a small subset of the data available in the agent's replay buffer $\B$ \citep{lin1993reinforcement}.\footnote{If the batch does not contain any transition that achieves a proto-goal, we are optimistic under uncertainty and classify it as plausible.}
Finally, we accept some latency by recomputing proxy values asynchronously, and only a few times  ($\approx 10$) per minute.
Section~\ref{sec:controllability_tests} shows that such a light-weight LSPI-based approach is indeed effective at identifying controllable goals.

\subsection{Desirability Weighting} 
\label{sec:desirability}

The second task of the PGE is to enable sampling the most \emph{desirable} goals from the reduced set of plausible goals $\G$ produced via pruning.
A lot has been discussed in the literature about what makes goals desirable \citep{gregor2016variational,bacon2017option,konidaris2009skill,eysenbach2018diversity,bellemare2016unifying,machado2017laplacian}; for simplicity, we stick to the two most commonly used metrics: novelty and reward-relevance. 
We use a simple count-based novelty metric \citep{auer2002using}:
\begin{equation}
    \label{eq:novelty}
    \nov(g) := \frac{1}{\sqrt{N(g)}},
\end{equation}
where $N(g)$ is the number of times goal $g$ has been achieved across the agent's lifetime.
The desirability score (or ``utility'') of a goal $g$ is then simply $u(g):=R(g) + \nov(g)$,
where $R(g)$ is the average extrinsic reward achieved on transitions where $g$ was achieved. Desirability scores for each goal are turned into a proportional sampling probability: 
\begin{equation}
    \P_\G(g) := \frac{u(g)}{\sum_{g'\in \G}u(g')}.
\end{equation}
In practice, when queried, the PGE does not output the full distribution, but a (small) discrete set of $K$ plausible and desirable goals, by sampling from $\P_\G$ with replacement ($K=100$ in all our experiments).

\section{Integrated RL Agent} 
\label{sec:gcrl}

This section details how to integrate proto-goal spaces and PGE components into a goal-conditioned RL agent.
As is typical in the goal-conditioned RL literature, we use a Universal Value Function Approximator (UVFA) neural network to parameterize the goal-conditioned action-value function $Q_{\theta}(s, a, g)$, which is eventually used to pick primitive actions.
At the high level, we note that the PGE is used at three separate \emph{entry points}, namely in determining how to act, what to learn about (in hindsight), and which goals to recombine.
What is shared across all three use-cases is the plausibility filtering of the goal space (implausible goals are never useful). 
However, the three use-cases have subtly different needs, and hence differ in the goal sampling probabilities.

\subsection{Which Goal to Pursue in the Current State} \label{sec:goal_selection}

For effective exploration, an agent should pursue goals that maximize its expected learning progress, i.e., it should pick a goal that will increase its competence the most \citep{lopes2012exploration}. 
As proxies for learning progress, we adopt two commonly used heuristics, namely novelty \citep{auer2002using} (Eq.~\ref{eq:novelty}) and reachability \citep{konidaris2009skill}. 
The issue with exclusively pursuing novelty is that this could lead the agent to prioritise the most difficult goals, which it cannot reach with its current policy yet, and hence induce behaviour that is unlikely to increase its competence. Thus, we combine novelty with a {\bf local reachability} metric, for which we can reuse the goal-conditioned value $V_{\theta}(s_t, g)$, which can be interpreted as the (discounted) probability that the agent can achieve goal $g$ from its current state $s_t$, under the current policy $\pi_{\theta}$.
To avoid computing reachability for each goal in $\G_t$ (which can be computationally expensive), we instead sample $M$ goals based on novelty and pick the closest: 
\begin{equation*}
g_t = \argmax_{g \in \{g_1, \ldots, g_M\} \sim \nov}\Big[V_{\theta}(s_t, g)\Big].
\end{equation*}

\subsubsection{Stratified Sampling over Heterogeneous Timescales}

The attainment count for a goal $N(g)$ can be low because it is rarely reachable, \textit{or} because it naturally takes a long time to reach.
To account for this heterogeneity in goal space, we first estimate each goal's natural timescale and then use \emph{stratified sampling} to preserve diversity and encourage apples-to-apples desirability comparisons. 
%
To estimate the characteristic timescale (or horizon) $h$ for each goal, we average the ``seek'' value-function over the state-space:
$    h(g) := \E_{s\sim\B}\Big[V_{\text{seek}}(s,g)\Big]$.
%
Once each goal has a timescale estimate, we divide the goals in the goal space into different buckets (quintiles). Then, we uniformly sample a bucket of goals; since the goals in the bucket have similar timescales ($\approx h$), we use novelty and reachability to sample a specific goal from that bucket to pursue (see Algorithm~\ref{alg:goal_selection_algorithm} in the appendix for details).

\subsubsection{Learning about Extrinsic Reward}
The evaluator always picks actions to maximize the extrinsic task reward. If the actors never did during training, then the action-value function would have unreliable estimates of the task return (called the \textit{tandem effect} \citep{georg_tandem_effect}). So, periodically, the actors pick the task reward function and select actions based on that. Since the task reward function may not correspond to a specific goal, we represent the task reward function as a special conditioning---a $\0$ tensor serves as the goal input to $Q_{\theta}(s, a, g=\0)$.

\subsection{Which Goals to Learn about in Hindsight} \label{sec:her}

Once the agent picks a goal $g$ to pursue, it samples a trajectory $\tau$ by rolling out the goal-conditioned policy $\pi_{\theta}(:, g)$. Given all the goals achieved in $\tau$, $\G^{\tau}_{A}$, the agent needs to decide which goals $G\subset\G^{\tau}_{A}$ to learn about in hindsight.

We always learn about the on-policy goal $g$, and the task reward (which corresponds to the conditioning $g=\0$). Among, the set of achieved goals $G\subset\G^{\tau}_{A}$, the agent samples a fixed set of $M_{her}$ goals and learns about them using hindsight experience replay \citep{andrychowicz2017neurips} (we use $M_{her}=15$).
Similar to the previous section, we want to sample those $M_{her}$ goals that maximize expected learning progress. We found that using a count-based novelty score as a proxy for learning progress (sample proportionally to $\nov(g)$) worked well for this purpose, and outperformed the strategies of (a) learning about all the achieved goals and (b) picking the $M_{her}$ goals uniformly at random from $\G^{\tau}_{A}$.

\begin{figure*}[tb]
    \centering
    \includegraphics[width=0.9\linewidth]{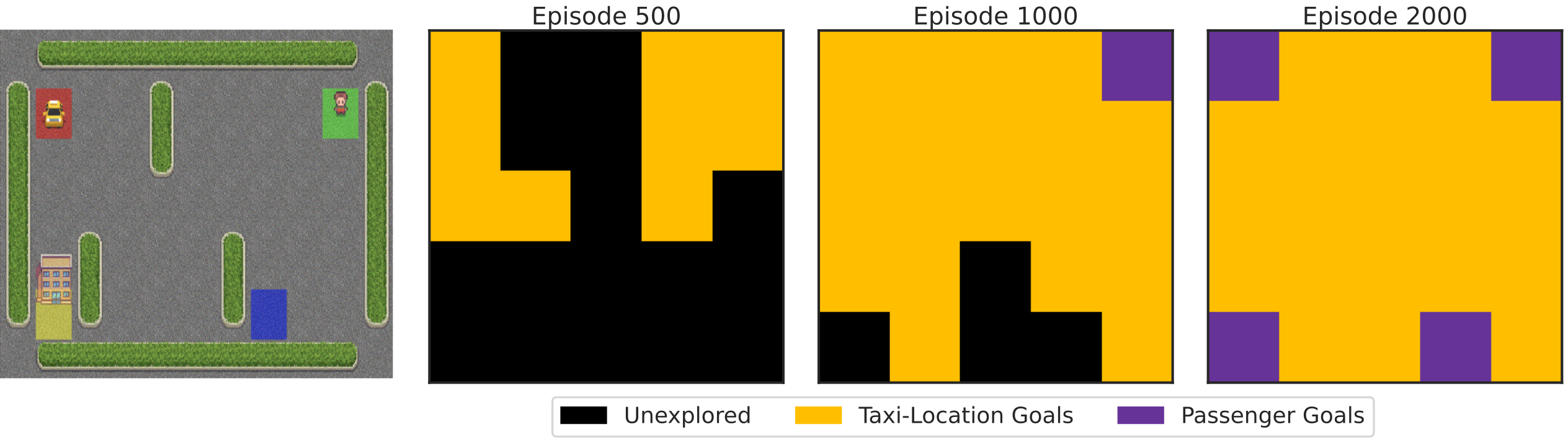}
    \vspace{-0.5em}
    \caption{Progression of the goal space refinement in \textsc{SparseTaxi} (domain illustrated in the leftmost sub-figure \citep{brockman2016openai}). This is a visualization of the $5\times5$ grid; yellow and black squares are explored and unexplored taxi locations respectively; purple squares denote explored passenger locations. The set of plausible goals grows over time from controlling the taxi location, to eventually controlling the location of the passenger. The passenger destination is always deemed uncontrollable by the Proto-goal Evaluator.}
    \vspace{-0.5em}
    \label{fig:goal_space_progression}
\end{figure*}

\begin{figure}[tb]
    \centering
    \includegraphics[width=\linewidth]{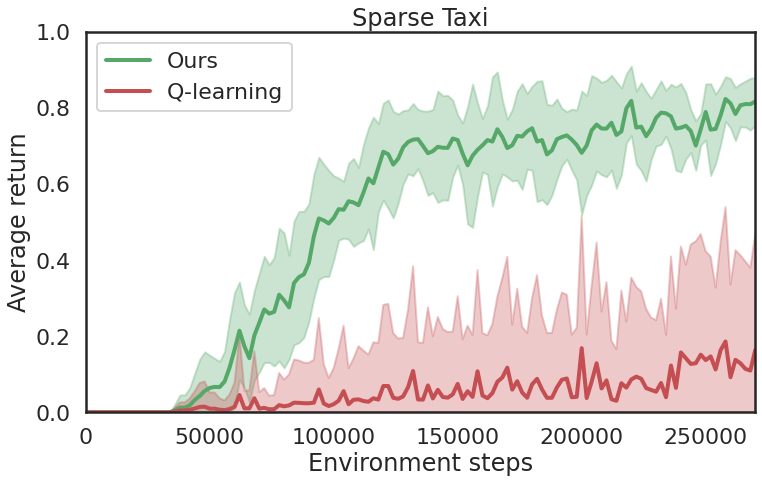}
    \vspace{-1.5em}
    \caption{Learning curves comparing Q-learning with $\epsilon$-greedy exploration to proto-goal-based exploration on \textsc{SparseTaxi}. Error bars denote standard error over $20$ independent runs.}
    \label{fig:taxi_lc}
\end{figure}

\subsection{Mastery-based Goal Recombination}
\label{sec:mastery}

We use one simple form of goal recombination in the agent: for any pair of goals that it has \emph{mastered}, it adds their combination (logical \texttt{AND}) as proto-goal candidate to be evaluated by the PGE.
A goal is considered mastered when its success rate is above a pre-specified threshold $\kappa$ ($=0.6$ in all our experiments).
For example, if the agent has mastered the goal of getting the key, and another goal of reaching the door, it will combine them to create a new proto-goal which is attained when the agent has reached the door with the key. Implementation details about creating and managing combination proto-goals can be found in the appendix (Algorithm~\ref{appendix:combination_pseudocode}).

\subsection{Distributed RL Agent Implementation}
For all non-toy experiments, we integrate our method with an off-the-shelf distributed RL agent, namely R2D2 \citep{kapturowski2018recurrent}.
It is a distributed system of $128$ actors asynchronously interacting with $128$ environments. The learner is Q-learning-based, using a goal-conditioned action-value function $Q_{\theta}(s, a, g)$ parameterized as a UVFA. Experience is stored in a replay buffer $\B$, including the binary proto-goal annotation vector $\bc$. More details about the agent, as well as pseudo-code, can be found in Appendix~\ref{sec:proto_r2d2_details}.

\section{Experiments}
\label{sec:experiments}

Our empirical results are designed to establish proto-goal RL an effective way to do exploration, first in a classic tabular set-up (\textsc{Taxi}, Section~\ref{sec:taxi}), and then at scale in two large-scale domains (\textsc{NetHack} and \textsc{Baba Is You}, Sections~\ref{sec:minihack} and~\ref{sec:baba}) whose combinatorial proto-goal spaces, left unpruned, would be too large for vanilla goal-conditioned RL.
Alongside, ablations and probe experiments show the effectiveness of our controllability and desirability metrics, and provide qualitative insights into the discovered goal spaces.

\subsection{Tabular Experiment: Exploration in \textsc{Taxi}}
\label{sec:taxi}

We build intuition about proto-goal exploration on the \textsc{Taxi} domain, a tabular MDP classically used to test hierarchical RL algorithms \citep{dietterich1998maxq}. In this problem, the agent controls a taxi in a $5\times5$ grid; the taxi must first navigate to the passenger, pick them up, take them to their destination (one of $4$) and then drop them off.
The default reward function is \emph{shaped} \citep{randlov1998learning}, 
but to make it a harder exploration problem, we propose the \textsc{SparseTaxi} variant, with two changes : (a) no shaping rewards for picking up or dropping off the passenger and (b) the episode terminates when the passenger is dropped off. In other words, the only (sparse) positive reward occurs when the passenger is dropped off at their correct destination. 

As proto-goal space, we use a factoring of the state space, namely one $b_i$ for each entity (taxi, passenger, destination) in each grid location ($|B|=34$).
Figure~\ref{fig:goal_space_progression} shows the progression of how the PGE gradually refines a goal space from those throughout training.
The set of reachable states expands gradually to mimic a curriculum; at first, goals correspond to navigating the taxi to different locations,  later they include goals for dropping off the passenger at different depots.
Also noteworthy is that proto-goals corresponding to the destination are absent from the goal space, because they are classified as uncontrollable. 

In terms of performance, our proposed method of goal-discovery also leads to more sample-efficient exploration in \textsc{SparseTaxi} (Figure~\ref{fig:taxi_lc}).
Compared to a vanilla Q-learning agent with $\epsilon$-greedy exploration, our goal-conditioned Q-learning agent learns to reliably and quickly solve the task.
More details can be found in Appendix~\ref{app:taxi}.

\begin{figure*}
    \centering
    \includegraphics[width=0.95\linewidth]{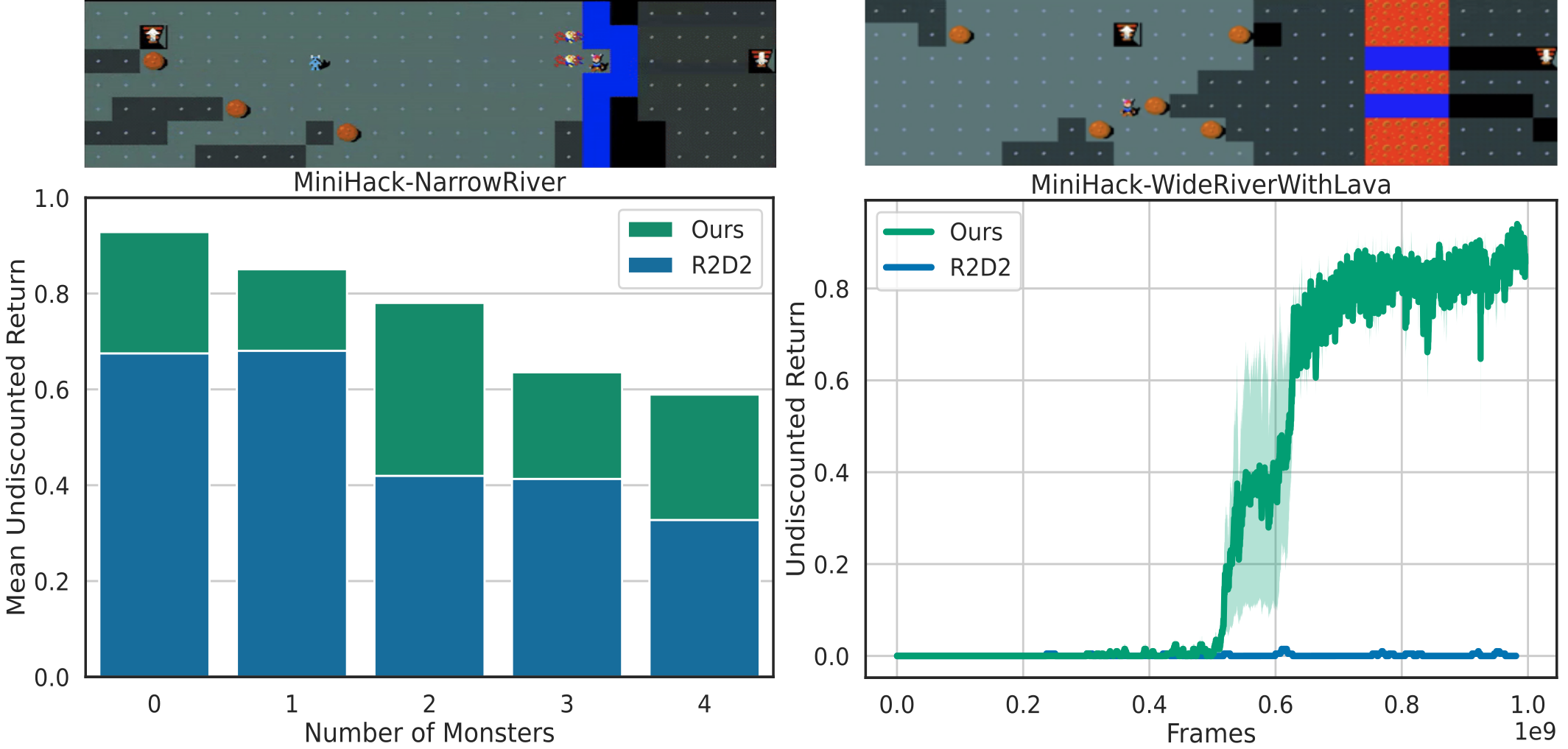}
    \vspace{-0.5em}
    \caption{\textsc{MiniHack} experiments. {\bf Left}: Final performance, when varying difficulty via the number of monsters in \textsc{NarrowRiver}. {\bf Right}: The challenge of getting off the ground in \textsc{WideRiverWithLava} (no monsters). 
    In all $5+1$ scenarios, using proto-goals outperforms the baseline R2D2 agent. Scenarios are illustrated above the result plots.}
    \vspace{-0.5em}
    \label{fig:minihack_results}
\end{figure*}

\subsection{Verifying the Controllability Measure} \label{sec:controllability_tests}

Our method of measuring controllability using the discrepancy between ``seek'' and ``avoid'' values (Section~\ref{sec:plausibility}) is novel, hence we conduct a set of sanity-checks to verify that it can capture controllability in all its guises.
Three toy experiments probe three separate types of controllability:
\begin{description}
    \item[Static controllability:] Proto-goals whose attainment status does not change. The passenger destination in \textsc{Taxi} is a good example of this kind of uncontrollability---while the destination changes \textit{between episodes}, there is no \textit{single transition} in the MDP in which it changes.
    \item[Time-based controllability:] Some problems have a \emph{timer} that increments, but is not controllable by the agent. 
    We check whether our controllability metric classifies such time-based proto-goals as plausible, using $4\times4$ gridworld with a timer that increments from $1$--$100$ (which is the $\max$ number of steps in an episode).
    \item[Distractor controllability:] More generally, parts of the observation that change independently of the agent's actions are distractors for the purpose of controllability. For this test, we use a visual gridworld, where one image channel corresponds to the controllable player, and the two other channels have pixels that light up uniformly at random \citep{gregor2016variational} (often referred to as a ``noisy TV'' \citep{schmidhuber2010formal,burdaLargeScaleStudy2019}).
\end{description}
For each of these toy setups, we compare our controllability predictions (Eq.~\ref{eq:control}) to ground-truth labels, and find it to correctly classify which proto-goals are controllable (Figure~\ref{fig:seek_avoid_results}, see Appendix~\ref{app:control} for details).
The prediction quality depends on the amount of data used to estimate the seek/avoid values.

\begin{figure*}[tb]
    \centering
    \vspace{-0.5em}
    \includegraphics[width=0.88\linewidth]{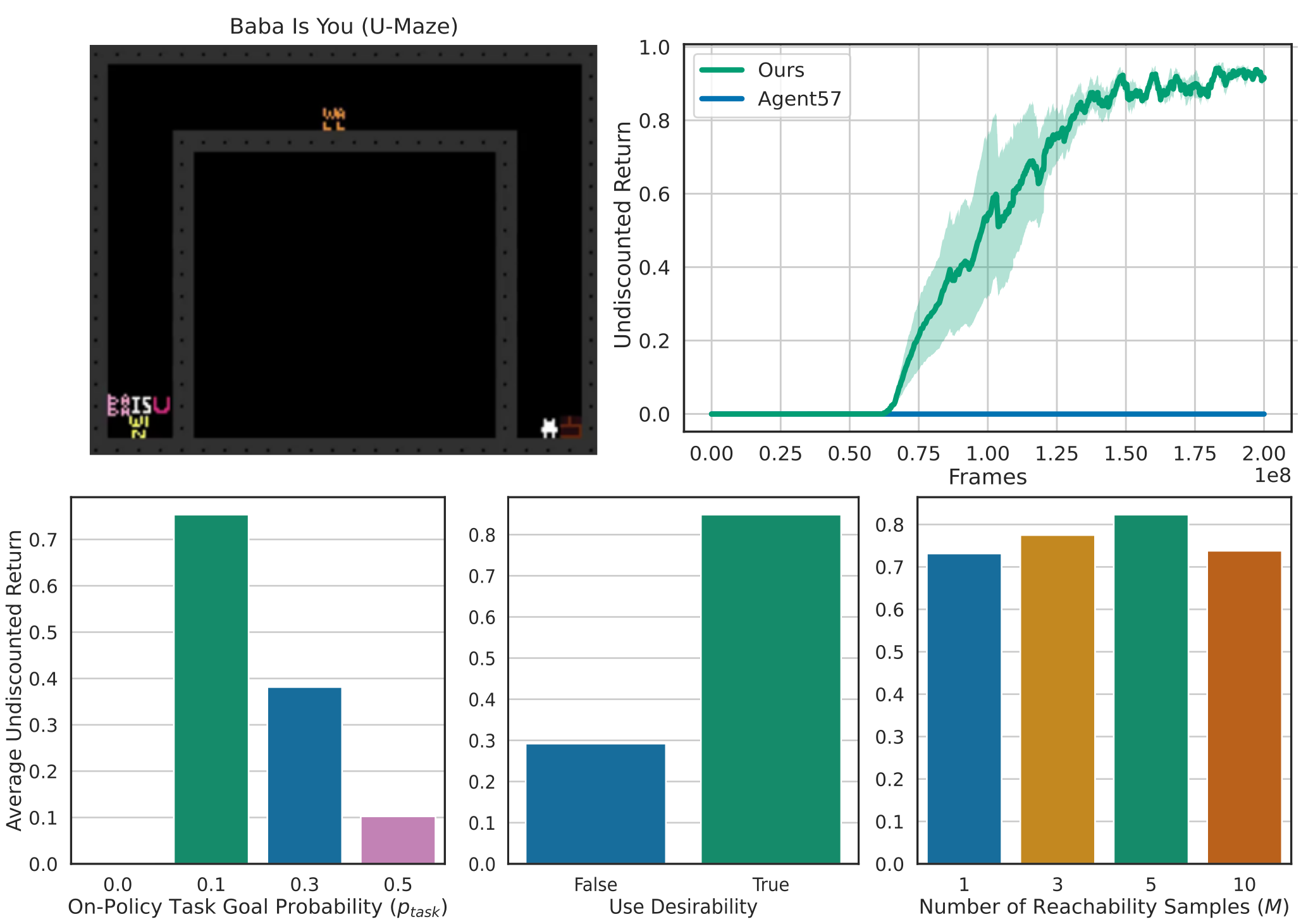}
    \caption{\textsc{Baba Is You} experiments. {\bf Top left}: \textsc{U-Maze} domain. {\bf Top right}: Learning curves comparing our approach to (flat-lining) Agent57. 
    {\bf Bottom}: Ablations showing the importance of sometimes acting according to the task reward during training \textit{(left)}, using desirability metrics in the PGE \textit{(middle)}, and the impact of the number of goals sampled for computing local reachability \textit{(right)}.
    }
    \vspace{-0.5em}
    \label{fig:baba_results}
\end{figure*}



\subsection{Natural Language Proto-goals: \textsc{MiniHack}}
\label{sec:minihack}

The first large-scale domain we investigate is 
\textsc{MiniHack} \citep{samvelyan2021minihack},
a set of puzzles based on the game \textsc{NetHack} \citep{kuttler2020nethack}, which is a grand challenge in RL. 
In addition to image-based observations, the game also provides natural language messages. This space of language prompts serves as our proto-goal space---while this space is very large (many 1000s of possible sentences), it contains a few useful and interesting goals that denote salient events for the task. Figure~\ref{fig:minihack_proto_illustration} illustrates how word-based proto-goal vectors are created in \textsc{MiniHack}.

We use two variants of the \textsc{River} task as exemplary sparse-reward exploration challenges. We choose them because \citet{samvelyan2021minihack}'s survey noted that while novelty-seeking algorithms \citep{burda2018exploration,raileanu2020ride} could solve the easiest version of \textsc{River}, they were unable to solve more difficult variations.

In all of these, the agent must make a bridge out of boulders and then cross it to reach its goal.
In the \textsc{NarrowRiver} variant, the agent needs to place one boulder to create a bridge, and the difficulty depends on the number of monsters who try to kill the player. 
Figure~\ref{fig:minihack_results} \textit{(left)} shows that while increasing the number of monsters degrades performance, our proto-goal agent outperforms the baseline R2D2 agent on each task setting.
In the \textsc{WideLavaRiver} variant, the river is wider, requiring $2$ boulders for a bridge, and includes deadly lava that also dissolves boulders.
Figure~\ref{fig:minihack_results} \textit{(right)} shows that our proto-goal agent comfortably outperforms its baseline.

\paragraph{Discovered goal space.} 
Words corresponding to important events in the game find their way into the goal-space. For instance, the word ``water'' appears in the message prompt when the boulder is pushed into the water and when the the player falls into the river and sinks. Later, combination goals like ``boulder'' \texttt{AND} ``water'' also appear in the goal-space and require the agent to drop the boulder into the water.

\subsection{Doubly Combinatorial Puzzles: \textsc{Baba Is You}}
\label{sec:baba}
The game \textsc{Baba Is You} \citep{teikari2019baba} has fascinating emergent complexity.
At first sight, the player avatar is a sheep (``Baba'') that can manipulate objects in a 2D space.
However, some objects are ``word-blocks'' that can be arranged into sentences, at which point those sentences become new rules that affect the dynamics of the game (e.g., change the win condition, make walls movable, or let the player control objects other than the sheep with an ``X-is-you''-style rule).
The natural reward is to reach the (sparse) win condition of the puzzle after $100$s of deliberate interactions. 

When testing various RL agents on \textsc{Baba is You}, we observed a common failure mode: the exploration process does not place enough emphasis on manipulating object and text blocks (see also Appendix~\ref{app:baba_details}). So, we created a simple level (\textsc{U-Maze} shown in Figure~\ref{fig:baba_results}) that is designed to focus on the crucial aspect of rule manipulation. This puzzle requires the agent to learn how to push a word block in place (from center to bottom left), which adds a new win-condition, and then touch the correct block (on the bottom right).
Exploration here is challenging because the agent has to master both navigation and block-manipulation before it can get any reward. In addition, the game's combinatorially large state space is a natural challenge to any novelty-based exploration scheme.

As in \textsc{Taxi}, we use a simple factored proto-goal space, with one binary element for every object (specific word blocks, wall, sheep) being present at any grid-position.
Plausible $1$-hot goals could target reaching a specific position of the sheep or movable blocks. 
Most combinations ($2$-hot proto-goals) are implausible, such as asking the sheep to be in two locations at once, but some could be useful, e.g., targeting particular positions for both ``Baba'' \textit{and} a word-block.

Given the exploration challenges in this domain (R2D2 never sees any reward, even on smaller variants of the puzzle), we use the stronger, state-of-the-art Agent57 agent as baseline here, which adds deep exploration on top of R2D2---it constructs an intrinsic reward using novelty and episodic memory  \citep{badia2020agent57}.
Figure~\ref{fig:baba_results} \textit{(top right)} shows that our R2D2 with proto-goals (but no intrinsic rewards) outperforms Agent57.
Note that with careful tuning, Agent57 does eventually get off the ground on this task, but never within the $200$M frame budget considered here (see Appendix~\ref{app:agent57} for details). On the other hand, Agent57 has the advantage that it does not require engineering a proto-goal space.

\paragraph{Discovered goal space.} 
At first, the goal-space is dominated by navigation goals; once these are mastered, goals that move the word-blocks begin to dominate. Then the agent masters moving to a particular location \textit{and} moving a word-block to some other location. Eventually, this kind of exploration leads to the agent solving the problem and experiencing the sparse task reward.

\paragraph{Ablations.} 
Figure~\ref{fig:baba_results} \textit{(bottom left)} analyzes how often the agent should act according to the extrinsic reward instead of picking a goal from the discovered goal-space. When that probability is $0$, the agent never reaches the goal during evaluation; acting according to the task reward function $10\%$ of the time during training performed the best in this setting.
In a second ablation, Figure~\ref{fig:baba_results} \textit{(bottom middle)} shows the importance of using desirability metrics on top of plausibility when mapping the proto-goal space to the goal-space.
Finally, Figure~\ref{fig:baba_results} \textit{(bottom right)} shows the impact of the number of goals sampled for computing local reachability during goal-selection (Section~\ref{sec:goal_selection}). 
Appendix~\ref{app:ablations} details other variants tried, how hyperparameters were tuned, etc.

\section{Conclusion and Future Work}
\label{sec:conclusion}

We presented a novel approach to using goal-conditioned RL for tackling hard exploration problems.
The central contribution is a method that efficiently reduces vast but meaningful proto-goal spaces to a smaller sets of useful goals,
using plausibility and desirability criteria based on controllability, reachability, novelty and reward-relevance.
Directions for future work include generalising our method to model-based RL to plan with jumpy goal-based models, more fine-grained control on when to switch goals \citep{pislar2021agents}, making the proto-goal space itself learnable, as well as meta-learning the ideal trade-offs between the various desirability criteria.

\section*{Acknowledgements}
\label{sec:acknowledgements}
We thank Alex Vitvitskyi for his patient help with the R2D2 code, and Steven Kapturowski and Patrick Pilarski (who spotted a double comma in a reference!) for feedback on an earlier draft. We also thank John Quan, Vivek Veeriah, Amol Mandhane, Dan Horgan, Jake Bruce and Charles Blundell for their support and guidance.


{
\bibliographystyle{named}
\bibliography{ijcai22}
}


\appendix

\section{\textsc{Taxi} Details} 
\label{app:taxi} 
In Section~\ref{sec:taxi}, we discussed our experiments with Proto-goal RL on the \textsc{Taxi} domain. Here, we share more details about the problem set up and the agent design.

\paragraph{Proto-goal space.} Each state $s_t$ is accompanied by a proto-goal vector $\mathbf{b}_t$, which is a $34$ dimensional binary vector: a $25$-bit $1$-hot vector describes the possible taxi locations, a $5$-bit $1$-hot vector describes the passenger location ($4$ depots and $1$ bit to determine whether the passenger is inside the taxi) and a $4$-bit $1$-hot vector describes the passenger's desired destination. These three $1$-hot vectors are concatenated to form the proto-goal vector $\bc_t$.

\paragraph{Baseline agent details.} 
The baseline Q-learning agent maintains a Q-table, $Q: \S \times \A \rightarrow \mathbb{R}$, which is initialized to zero. During training, the agent does $\epsilon$-greedy exploration with $\epsilon=0.1$. Periodically, the agent is tested by rolling out the current greedy policy $\pi=\arg\max_a Q(s, a)$ with $\epsilon=0$.

\paragraph{Proto-goal agent details.} 
Algorithm $\ref{alg:tabular_pgrl}$ describes the proto-goal RL agent used for our experiments in tabular \textsc{taxi}. The agent maintains two different Q-tables for each proto-goal $Q_{\text{seek}}^g$ and $Q_{\text{avoid}}^g$; these Q-tables are first used by the proto-goal evaluator to determine the goal-space $\G$ and subsequently to pick actions conditioned on the goal $g\sim G$ picked for the current episode. All transitions $(s, a, r, s')$ are used to update all proto-goal conditioned Q-functions. In addition to learning goal-conditioned Q-functions, the Proto-goal agent maintains one Q-table for the extrinsic reward function $Q(s,a) = \E[R_1+R_2+\ldots|s_t=s,a_t=a]$., which is used to greedily pick actions during evaluation.

\begin{algorithm}[tb]
    \caption{Tabular Proto-goal Agent}
    \label{alg:tabular_pgrl}
    \begin{algorithmic}[1] 
        \STATE Initialize task Q-table: $Q(s,a)=0,\forall{s\in\S,a\in\A}.$
        \STATE Initialize proto-goal Q-tables: $$Q^g_{\text{seek}}(s,a)=Q^g_{\text{avoid}}(s,a)=0,\forall{s\in\S,a\in\A,g\in B}.$$
        \WHILE{training}
            \STATE Determine goal space: $\G_t$=\texttt{pge}($B_t$)
            \STATE Pick a goal $g=$\texttt{select\_goal}($s_t,\G_t$)
            \WHILE {not \texttt{done} and not $g$ achieved in $s_t$}
                \STATE $a_t = \arg\max_{a\in\A}Q^g_{\text{seek}}(s_t, a)$
                \STATE $R_t, s_{t+1},\bc_t,$ \texttt{done} = \texttt{env.step}($s_t,a_t$)
                \STATE Add proto-goals $\bc_t$ to the list of all proto-goals $B_t$.
                \STATE Update the task Q-function using TD-error $\delta_t$ $$\delta_t=R_t+\gamma\max_a Q(s_{t+1}, a)-Q(s_t,a_t).$$
                \STATE Update the proto-goal seek Q-functions $\forall g\in B_t$ $$\delta^g_t=r^g_{\text{seek}}+\gamma\max_a Q_{\text{seek}}^g(s_{t+1}, a)-Q_{\text{seek}}^g(s_t,a_t).$$
                \STATE Update the proto-goal avoid Q-functions $\forall g\in B_t$ $$\delta^g_t=r_{\text{avoid}}^g+\gamma\max_a Q_{\text{avoid}}^g(s_{t+1}, a)-Q_{\text{avoid}}^g(s_t,a_t).$$
            \ENDWHILE
        \ENDWHILE
    \end{algorithmic}
\end{algorithm}

\begin{algorithm}[tb]
    \caption{Actor::\texttt{select\_goal()}}
    \label{alg:goal_selection_algorithm}
    \textbf{Inputs}: Goal space $\G$, current state $s_t$\\
    \textbf{Output}: Goal to pursue at $s_t$\\
    \textbf{Hyperparameters}: Number of timescale bins $k$, number of novel goals $m$, task reward probability $p_{tas k}$. 
    \begin{algorithmic}[1] 
        \IF{\texttt{random.random()}$<p_{task}$}
            \STATE \textbf{return} $g_t=\0$
        \ELSE
            \STATE Bin goals $\G=\{\G_1,...,\G_k\}$ based on timescale.
            \STATE Uniformly sample a bin from $\G$: $\G^h=\{g_1,...,g_n\}$.
            \STATE Sample (with replacement) a set $G=\{g_1,...,g_m\}$ from $\G^h$ based on novelty.
            \STATE Pick nearest goal in $G$: $g_t=\arg\max_{g\in G}\Big[V_{\theta}(s_t|g)\Big]$
            \STATE \textbf{return} $g_t$
        \ENDIF
    \end{algorithmic}
\end{algorithm}

\begin{algorithm}[tb]
    \caption{PGE::\texttt{get\_plausible\_proto\_goals()}}
    \label{alg:pge_plausibility}
    \textbf{Inputs}: Proto-goals observed so far $B_t$, replay buffer $\B$.\\
    \textbf{Hyperparameters}: Controllability threshold $\tau_1$, reachability threshold $\tau_2$. 
    \begin{algorithmic}[1] 
        \STATE Sample transitions from replay: $s,a,s'\sim\B$.
        \STATE Relabel transitions based on seek reward function: $r^g_{\text{seek}}=R_{\text{seek}}(s,s',g)$, $\forall{g}\in B_t$.
        \STATE Relabel transitions based on avoid reward function: $r^g_{\text{avoid}}=R_{\text{avoid}}(s,s',g)$, $\forall{g}\in B_t$.
        \STATE Train seek value-function for each proto-goal: $V_{\text{seek}}^g=$\texttt{LSPI}($s,a,r^g_{\text{seek}},s'$), $\forall{g}\in B_t$.
        \STATE Train avoid value-function for each proto-goal: $V_{\text{avoid}}^g=$\texttt{LSPI}($s,a,r^g_{\text{avoid}},s'$), $\forall{g}\in B_t$.
        \STATE Get the controllable proto-goals: $$\mathcal{C}=\Big\{g: \E_s\Big[V_{\text{seek}}^g(s)+V_{\text{avoid}}^g(s)\Big] < \tau_1, \forall{g}\in B_t\Big\}.$$
        \STATE Plausible ones are the sometimes-reachable subset: $$\mathcal{G}=\Big\{g:
        \max_s V^g_{\text{seek}}(s)>\tau_2,\forall{g}\in \mathcal{C}\Big\}.$$
        \STATE \textbf{return} plausible proto-goals $\mathcal{G}$.
    \end{algorithmic}
\end{algorithm}

\begin{algorithm}[tb]
    \caption{PGE::\texttt{get\_desirable\_goals()}}
    \label{alg:pge_desirability}
    \textbf{Inputs}: Plausible proto-goals $\G$,  mapping of proto-goals to counts $N: \G\rightarrow \mathbb{N}$,  mapping of proto-goals to average reward $R: \G\rightarrow \mathbb{R}$.\\
    \textbf{Hyperparameters}: Size of goal set $K$.\\ 
    \begin{algorithmic}[1] 
        \STATE Assign each proto-goal a desirability score: $$u(g)=R(g)+\frac{1}{\sqrt{N(g)}}, \forall{g\in \G}.$$
        \STATE Convert the desirability score to probabilities: $$\P(g)=\frac{u(g)}{\sum_g u(g)}, \forall{g\in\G}.$$
        \STATE Sample with replacement $G$, a set of $K$ goals from plausible proto-goals $\G$ using desirability probabilities $\P(g)$.
        \STATE \textbf{return} goal set $G$.
    \end{algorithmic}
\end{algorithm}

\begin{algorithm}[tb]
    \caption{PGE::\texttt{get\_combination\_proto\_goals()}}
    \label{appendix:combination_pseudocode}
    \textbf{Inputs}: Goal space $\G$, mapping of goals to counts $N: \G \rightarrow \mathbb{N}$, mapping of goals to the average success ratio on the last $10$ on-policy pursuits $S: \G \rightarrow [0,1]$.\\
    \textbf{Hyperparameters}: Mastery threshold $\kappa$.\\ 
    \begin{algorithmic}[1] 
        \STATE Mastered goals $M$: $$M=\{g|g\in\G, N(g)\geq 10,S(g)>\kappa\}$$
        \STATE Convert the mastery scores to probabilities: $$\P(g)=\frac{S(g)}{\sum_g S(g)}, \forall{g\in M}.$$
        
        \STATE Sample (without replacement) $2$ goals from $\P(g): g_1, g_2$.
        \STATE Create new proto-goal $\bc=$ \texttt{AND}($g_1,g_2$) and add it to the proto-goal space $B_t$.
        \STATE \textbf{return} $B_t$
    \end{algorithmic}
\end{algorithm}

\begin{algorithm}[t]
    \caption{Actor::\texttt{run\_loop()}}
    \label{alg:actor_run_loop}
    \textbf{Inputs}: Proto-goal evaluator \texttt{pge}.\\
    \textbf{Hyperparameters}: Number of goals to replay $M_{her}$.\\ 
    \begin{algorithmic}[1] 
        \WHILE{\texttt{true}}
            \STATE $g=$\texttt{select\_goal}($s_t$).
            \STATE Sample trajectory $\tau_g=\{s,a,r_g,s',b\}$ by rolling out goal-conditioned policy $\pi_{\theta}(s_t|g).$ \label{line:sample_pi_g}
            \STATE Add $\tau_g$ to the replay buffer.
            \STATE Get $\G_A$, the set of goals achieved during $\tau_g$.
            \STATE Sample $\G_A^\text{novel}$ without replacement from $\G_A$, a set of $\min(M_{her}, |\G_A|)$ goals,
            proportionally  to $\frac{1}{\sqrt{N(g)}}$.
            \STATE Relabel trajectory $\tau_g$, once for each goal in $\G_A^\text{novel}$.
            \STATE Relabel trajectory $\tau_g$ with extrinsic reward.
            \STATE Add relabeled trajectories to the replay buffer.
            \STATE Use $\tau_g$ to update the proto-goal buffer $B_t$, their counts and their average reward.
        \ENDWHILE
    \end{algorithmic}
\end{algorithm}

\section{Details of Controllability Experiments}
\label{app:control}
\begin{figure*}[t]
    \centering
    \includegraphics[width=0.95\linewidth]{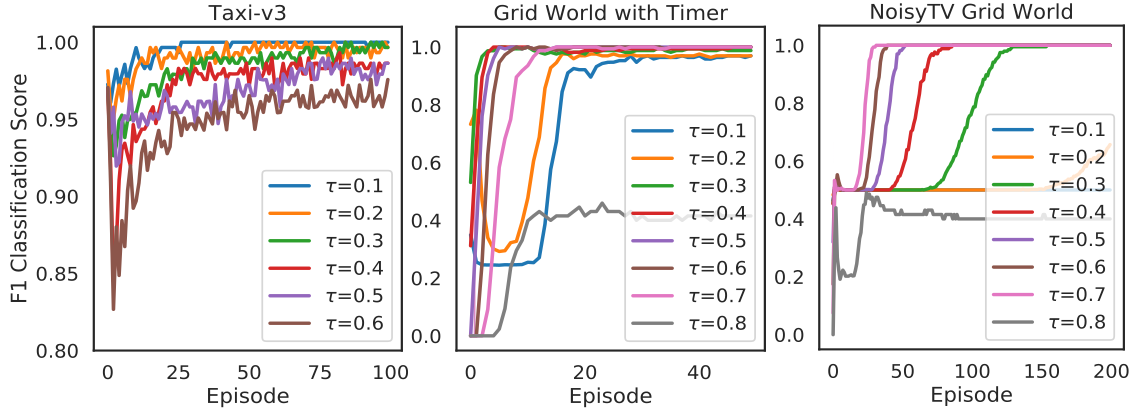}
    \caption{Testing our seek-avoid controllability measure in toy domains using the F1 classification metric. In each domain, we collect transitions using a random policy and measure how effectively we can classify controllability for different values of the threshold $\tau_1$.}
    \label{fig:seek_avoid_results}
\end{figure*}
In each of the toy problem setups described in Section~\ref{sec:controllability_tests} of the main paper, we roll out a random policy and store the collected transitions in memory. At the end of each episode, we re-train the seek- and avoid-value functions on all transitions, as discussed in Section~\ref{sec:seek_avoid_lspi}, and use Eq.~\ref{eq:control} to classify whether or not different proto-goals are controllable. We then test the predicted classification labels against the ground-truth labels using the F1-classification score.
\paragraph{Results.} Figure~\ref{fig:seek_avoid_results}  shows we are able to correctly classify proto-goals as controllable vs uncontrollable in each of the aforementioned cases. These results also suggest that training the seek-avoid value functions using more data improves classification performance.

\section{Details of the Proto-R2D2 Agent} \label{sec:proto_r2d2_details}
\label{app:uvfa_architecture}

\begin{figure}[tb]
    \centering
    \includegraphics[width=\linewidth]{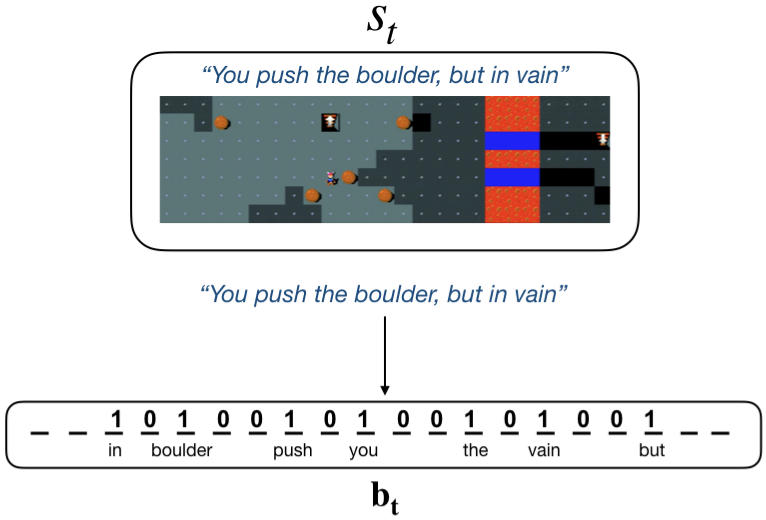}
    \vspace{-0.5cm}
    \caption{Proto-goal space in \textsc{MiniHack}: the message part of the observation is converted to a binary vector $b_t$ using bag-of-words.}
    \label{fig:minihack_proto_illustration}
\end{figure}

\begin{figure}[tb]
    \centering
    \includegraphics[width=\linewidth]{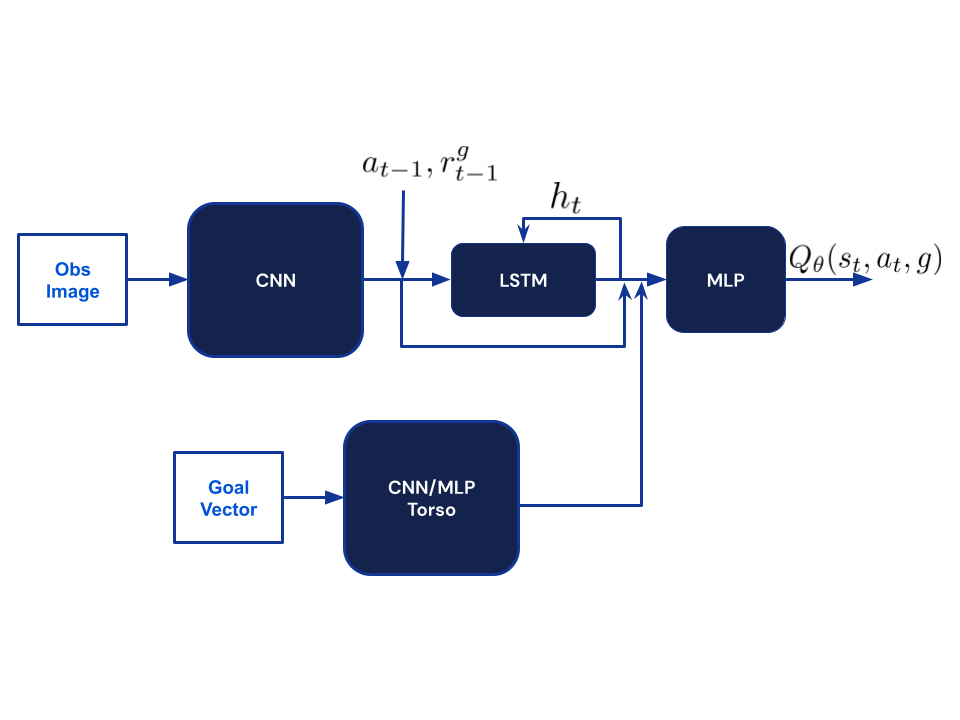}
    \vspace{-1.75cm}
    \caption{UVFA Architecture used to learn $Q_{\theta}(s, a, g)$. }
    \label{fig:uvfa_architecture}
\end{figure}

\begin{figure}[t]
    \centering
    \includegraphics[width=\linewidth]{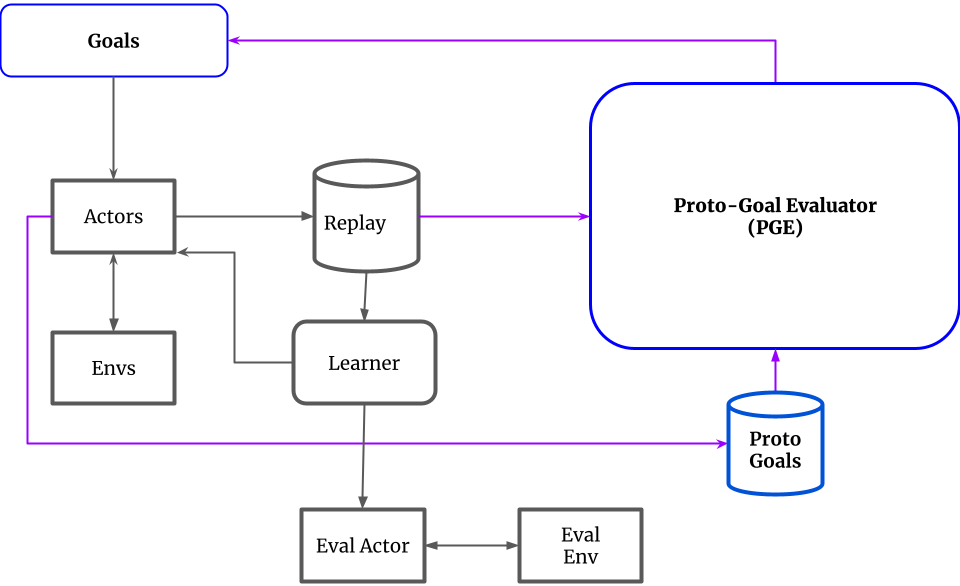}
    \caption{System architecture of the Proto-goal R2D2 Agent. Gray blocks denote components shared with baseline R2D2, blue blocks denote components added for Proto-goal RL.}
    \label{fig:sys_architecture}
\end{figure}

We now describe the details of the agent used to produce the results in Sections~\ref{sec:minihack} and~\ref{sec:baba}. The R2D2 algorithm \citep{kapturowski2018recurrent} forms the basis for the Proto-R2D2 agent; so, there are $128$ actors interacting with $128$ copies of the environment. 

These actors follow the process described in Algorithm~\ref{alg:actor_run_loop}. The goal-conditioned value-function (UVFA) used to select actions in line \ref{line:sample_pi_g} is parameterized using the network architecture shown in Figure~\ref{fig:uvfa_architecture}. This UVFA is trained using R2D2's version of Q-learning on batches randomly sampled from the replay buffer $\{s, a, r_g, s', g\}\sim\B$. As shown in Figure~\ref{fig:sys_architecture}, each of the actors stream the proto-goals that they observe to the proto-goal evaluator (PGE), which maintains the buffer $B_t$ of ever-seen proto-goals, with the number of times each proto-goal has been achieved $N(g)$. Each actor also streams the extrinsic reward observed when each proto-goal was achieved; this information is also streamed to the PGE, which keeps a running mean of the rewards.

The PGE asynchronously samples from replay and determines the goal-space for goal-conditioned RL. It first prunes the proto-goal buffer $B_t$ using plausibility metrics as described in Algorithm~\ref{alg:pge_plausibility}. Then, it samples a smaller space of goals using desirability metrics as described in Algorithm~\ref{alg:pge_desirability}. This goal-space is subsequently used in the next actor/learner iteration.

The evaluator is another process that asynchronously interacts with its own copy of the environment; the rewards it experiences are reported in the learning curves in Figures~\ref{fig:minihack_results} and~\ref{fig:baba_results} (main paper). The evaluator also picks actions using the UVFA parameters $\theta$, but it always conditions the UVFA on task reward-function (using $g=\0$ as discussed in the main paper). Following \citet{badia2020agent57}, all our learning curves are averaged over $3$ random seeds.

\subsection{Agent57 Baseline}
\label{app:agent57}

We performed a grid search over Agent57 hyperparameters \texttt{episodic\_memory\_reward\_scale}$=\{0, 0.01, 0.001\}$ and \texttt{meta\_episode\_length}$=\{1, 5, 10\}$ on the \textsc{U-Maze} task. \texttt{meta\_episode\_length} represents the number of episodes after which the episodic memory is reset, \texttt{episodic\_memory\_reward\_scale} modulates the contribution of the episodic memory to the intrinsic reward \citep{badia2020agent57}. Out of these, \texttt{episodic\_memory\_reward\_scale} of $0.001$ and \texttt{meta\_episode\_length} of $10$ performed the best; the results presented in the paper use these hyperparameters. None of these configurations get off the ground in the $200$ million frames range considered here; however, we have evidence that, when trained up to $1$ \textit{billion} frames, Agent57 with the chosen hyperparameters can solve \textsc{U-Maze}.

\section{More Details about Test Environments}
In this section, we outline some more details about the set up for our \textsc{MiniHack} and \textsc{Baba Is You} experiments. 

\subsection{\textsc{MiniHack}}
\label{app:minihack_details}

The puzzle is procedurally generated every episode. The action space in our tasks was all $8$ compass directions. The tasks are partially observable---not all tiles are visible to the agent at all times; the player must go close to a tile to reveal its contents. Even when the number of monsters is configured to $0$, because of the dynamics of \textsc{NetHack}, there is always \textit{some} probability that a monster will spawn in a tile. Episodes last for a maximum of $350$ steps, unless the player reaches the goal, in which case the episode terminates with a sparse reward of $1$.

\subsection{\textsc{Baba Is You}}
\label{app:baba_details}

We use a DMLab \citep{beattie2016deepmind} implementation of the popular game \textsc{Baba Is You} \citep{teikari2019baba}. The observation is a stack of 2D images: each image in the stack corresponds to a single object. For example, one image plane represents the sheep \textit{baba}, while another image plane represents the word-block ``baba''. Since there are $7$ objects in the \textsc{U-Maze} level, each observation is a $2$D image with $7$ channels. There are $5$ actions in the action-space: up, down, left, right and no-op. Episodes last for a maximum of $1000$ steps. If there is no rule that specifies that \texttt{X-IS-YOU} (where \texttt{X} is any object), then the episode terminates with a reward of $0$ (because there is nothing left to control); if the agent reaches whatever object forms the win condition (as in \texttt{Y-IS-WIN}), then the episode terminates with a sparse reward of $1$. 

The preliminary experiments that let to the design to the \textsc{U-Maze} used various RL agents, including V-trace-based planning agents, R2D2, Q-learning, Agent57, tested on different (early) in-game levels of \textsc{Baba is You}.


\section{Further Ablations and Variants Tried}
\label{app:ablations}

In this section, we discuss some of the strategies that did \textit{not} work and some strategies we used to set the values of certain hyperparameters.

\begin{enumerate}
    \item Importance of the timescale-based stratification: without timescale stratification (Section~\ref{sec:goal_selection}), the local reachability metric always hurt performance, likely because it biased goal-selection towards easy goals. After incorporating stratified sampling, using local reachability outperformed only using novelty for goal-selection and allowed the agent to condition the higher-level policy on the current state.
    \item Number of hindsight goals replayed: Each trajectory is used to replay a maximum of $M_{her}=15$ achieved goals in hindsight (Section~\ref{sec:her}). We swept over this value: performance tends to follow the familiar U-shaped curve---replaying too few or too many goals degrades performance. 
    \item Importance of the plausibility metrics: without the use of the plausibility heuristics discussed in Section~\ref{sec:plausibility}, the system drowns in goals it can never achieve. For example, the vast majority of combination goals are implausible and yet they are very novel; without pruning, these goals dominate the goal-space and prevent true competence progress.
\end{enumerate}

\begin{table}[tb]
\centering
\begin{tabular}{l|c}
\toprule
\textbf{Parameter} & \textbf{Value}\\
\midrule
Replay buffer size & $1e6$  \\
Prioritized experience replay$^*$  & False \\
Optimizer & Adam \\
Learning rate & $3 \cdot 10^{-4}$\\
Adam hyper-parameters & $\beta_1=0.9$\\
  &  $\beta_2=0.999$\\
  & $\epsilon=10^{-4}$\\
Weight decay & $10^{-4}$\\
Target net update period & $1e4$  \\
Batch size & 64  \\
Discount factor $\gamma$ & 0.99 \\
Trace length$^*$ & 10 \\
Trace $\lambda$ & 0.95 \\
\bottomrule
\end{tabular}
\caption{Inherited R2D2 Hyperparameters. These are essentially untuned except for those marked with $^*$ (see Section~\ref{app:hypers}).
}
\label{tab:r2d2_hyperparameters}
\end{table}

\begin{table}[tb]
\centering
\begin{tabular}{l|c|c}
\toprule
\textbf{Parameter} & \textbf{Value} & \textbf{Tuned?} \\
\midrule
Sampled goal set size $K$ & $100$ & no\\
Number of goals to replay $M_{her}$ & $15$ & yes\\
Controllability threshold $\tau_1$ & $0.1$ & yes\\
Reachability threshold $\tau_2$ & $0.5$ & no\\
Number of random projections $|\phi|$ & 32 & no\\ 
LSPI batch size & $1024$ & no\\
LSPI discount factor & $0.95$ & no\\
Task reward probability $p_{task}$ & $0.1$ & yes\\
Number of timescale buckets $k$ & $5$ & no\\
Number of novelty goals to sample $M$ & $5$ & no\\
Mastery threshold $\kappa$ & $0.6$& yes\\
Number of seeds & 3 & N/A\\
\bottomrule
\end{tabular}
\caption{Specific Proto-R2D2 hyperparameters.}
\label{tab:proto_r2d2_hyperparameters}
\end{table}

\section{Hyperparameters}
\label{app:hypers}

Table~\ref{tab:r2d2_hyperparameters} lists the parameter settings for the R2D2 agent; Table~\ref{tab:proto_r2d2_hyperparameters} shows the settings for the hyperparameters specifically added by our algorithm. Here is the rationale for picking the values of some of the hyperparameters:
\begin{enumerate}
    \item LSPI batch size: this is the batch size for computing the seek/avoid value-functions. We used $1024$ because that was the largest batch size for which the wall-clock time for training was still reasonable.
    \item Number of random projections: we followed \citet{ghavamzadeh2010lstd} and set this to the square-root of the number of transitions in the minibatch. Since this heuristic worked well, we did not tune the number of random projections any further.
    \item Controllability threshold: we used $0.1$ because it worked well in smaller domains in which we could measure the F1 classification score against the ground-truth controllable goals (see Figure~\ref{fig:seek_avoid_results}).
    \item Number of timescale buckets: we manually looked at goal-spaces with different values of $k$. When $k=5$, the goal-space divided into goal buckets that intuitively segmented based on increasing difficulty.
    \item Trace length: R2D2 typically uses a larger trace length \citep{kapturowski2018recurrent}. Unlike Atari games, our tasks can have much shorter episode lengths (due to goal terminations), so we used a shorter trace length.
    \item Prioritization: we did not use prioritized experience replay (PER) \citep{schaul2015prioritized} because it would significantly increase computation and implementation complexity. More specifically, to replay a trajectory with an off-policy goal, we would have to recompute TD errors with respect the hindsight goal's reward function. Due to the added computation and implementation complexity (and because this was not our core contribution), we did not use prioritization.
\end{enumerate}

\paragraph{Network architecture.} 
Figure~\ref{fig:uvfa_architecture} shows the network architecture for the goal-conditioned Q-function. The network architecture is identical to R2D2 \citep{kapturowski2018recurrent} except for the fact that the goal is processed by its own CNN (when it is an image as in Baba Is You) or by an MLP (when the goal is a flattened binary vector as in MiniHack); the architecture of the goal-processing CNN/MLP is identical to the corresponding blocks in R2D2.

\paragraph{Hyperparameter sensitivity.} 
In general, we found that the proto-goal agent was robust to different settings of the majority of hyperparamters in Table~\ref{tab:proto_r2d2_hyperparameters}, with the exception of the task goal probability $p_{task}$, whose different settings resulted in qualitatively different performances (see Figure~\ref{fig:baba_results}).

\end{document}